\documentclass[final]{cvpr}

\usepackage{times}
\usepackage{epsfig}
\usepackage{graphicx}
\usepackage{amsmath}
\usepackage{amssymb}
\usepackage{comment}

\usepackage{multirow}
\usepackage{bm}
\usepackage{subfigure}
\newcommand{\tabincell}[2]{\begin{tabular}{@{}#1@{}}#2\end{tabular}}

\usepackage[pagebackref=true,breaklinks=true,colorlinks,bookmarks=false]{hyperref}

\begin{document}
\hyphenpenalty=1000

\title{Face Forgery Detection by 3D Decomposition}

\author{Xiangyu Zhu$^{1,2}$\thanks{Equal contribution.} \and Hao Wang$^{1*}$ \and Hongyan Fei$^{3}$ \and Zhen Lei$^{1,2}$\thanks{Corresponding author.} \and Stan Z. Li$^{4}$ \\
$^{1}$CBSR \& NLPR, Institute of Automation, Chinese Academy of Sciences,\\
$^{2}$School of Artificial Intelligence, University of Chinese Academy of Sciences,\\
$^{3}$School of Automation and Electrical Engineering, University of Science and Technology Beijing,\\
$^{4}$School of Engineering, Westlake University \\
{\tt \small \{xiangyu.zhu,zlei,szli\}@nlpr.ia.ac.cn} ~~ {\tt \small \{haowang7308,hongyanfei0420\}@gmail.com}
}

\maketitle

\begin{abstract}
    Detecting digital face manipulation has attracted extensive attention due to fake media's potential harms to the public.
    However, recent advances have been able to reduce the forgery signals to a low magnitude.
    Decomposition, which reversibly decomposes an image into several constituent elements, is a promising way to highlight the hidden forgery details.
    In this paper, we consider a face image as the production of the intervention of the underlying 3D geometry and the lighting environment, and decompose it in a computer graphics view.
    Specifically, by disentangling the face image into 3D shape, common texture, identity texture, ambient light, and direct light, we find the devil lies in the direct light and the identity texture.
    Based on this observation, we propose to utilize facial detail, which is the combination of direct light and identity texture, as the clue to detect the subtle forgery patterns.
    Besides, we highlight the manipulated region with a supervised attention mechanism and introduce a two-stream structure to exploit both face image and facial detail together as a multi-modality task.
    Extensive experiments indicate the effectiveness of the extra features extracted from the facial detail, and our method achieves the state-of-the-art performance.
\end{abstract}

\section{Introduction}
While earlier seamless face manipulation has amazed the public broadly, there has been a constant concern about the potential abuse of relevant techniques.
In particular, the recent \emph{DeepFake}~\cite{deepfake2020github} initiated the widespread public discussion among the potential harmful consequence~\cite{deepfake_bbc_business} and feasible detection solutions of counterfeit facial media~\cite{deepfake_bbc_bitesize}.

\begin{figure}[htbp]
    \subfigure{
        \label{fig:lambertian-a}}
    \subfigure{
        \label{fig:lambertian-b}}
    \subfigure{
        \label{fig:lambertian-c}}
    \subfigure{
        \label{fig:lambertian-d}}
    \subfigure{
        \label{fig:lambertian-e}}
    \subfigure{
        \label{fig:lambertian-f}}
    \begin{center}
        \includegraphics[width=0.98\linewidth]{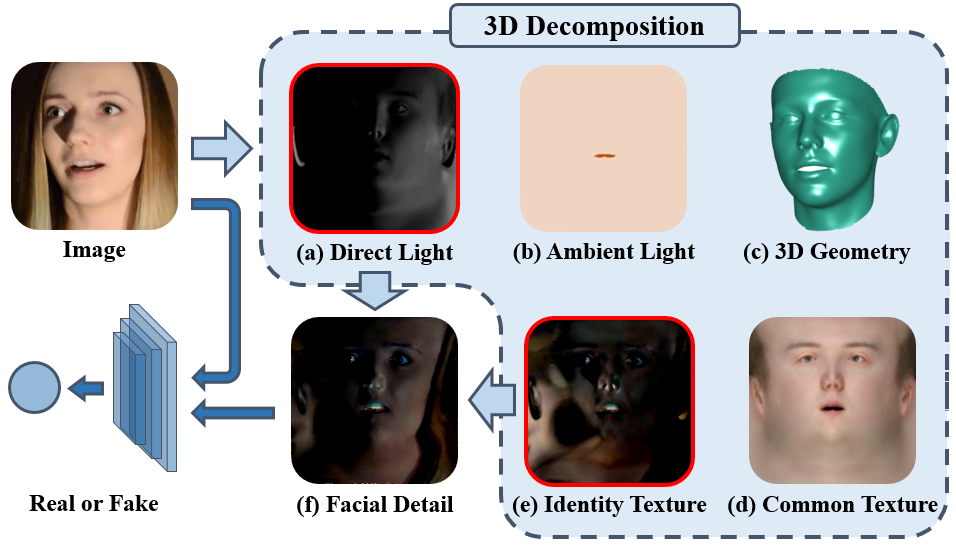}
    \end{center}
    \caption{In computer graphics, a face image can be decomposed into direct light, ambient light, 3D geometry, common texture and identity texture. We find direct light and identity texture contain critical clues and merge them as the facial detail for forgery detection.}
    \label{fig:lambertian}
\end{figure}

In this work, we are dedicated to detecting the manipulation on facial identity and expression, related to the very popular \emph{DeepFakes} (DF)~\cite{deepfake2020github}, \emph{Face2Face} (F2F)~\cite{thies2016face2face}, \emph{FaceSwap} (FS)~\cite{faceswap2020github} and \emph{NeuralTextures} (NT)~\cite{thies2019deferred}, which perform the state-of-the-art face manipulation, making it extremely tough to reveal the sophisticated counterfeit flaws from the image view only~\cite{qian2020thinking}.
This situation stimulates researchers to shift their attention to extracting forgery evidence from other aspects besides the original RGB image.

Previous work~\cite{zhang2019detecting,chen2020manipulated,wang2020cnn,qian2020thinking} has discovered that the signals in specific frequency ranges are replaced by particular patterns during manipulation and proposes to detect forgery by signal decomposition.
The assumption is that, by disentangling the face image, we can find more critical clues for forgery detection from the constituent elements, which are overlooked or hard to be forged by the manipulation methods, whose loss function mainly constrains pixel values.
For example, Zhang \etal~\cite{zhang2019detecting} identify the unique replications of spectra in the frequency domain due to the up-sampling process.
Chen \etal~\cite{chen2020manipulated} introduce facial semantic segmentation and \emph{Discrete Fourier Transform} (DFT) to extract both spatial- and frequency-domain features, respectively.
However, it is difficult to decide which range of signals contains artifacts since images are captured by different devices, under different environments, and even compressed with different algorithms, leading to large frequency distribution bias across datasets.
The hand-crafted~\cite{stuchi2017improving} and learned~\cite{qian2020thinking} frequency filters also easily suffer from the generalization problem.
Therefore, the crucial problems of this topic lie in how to decompose an image and how to identify reliable constituent elements.

In this paper, we consider the decomposition from a physics view that a face image is the intervention result of the underlying 3D geometry, its albedo, and the environment lighting.
Specifically, we introduce \emph{3D Morphable Model} (3DMM)~\cite{blanz2003face} and the \emph{computer graphics} rendering to simulate the generation of a face image.
Under \emph{Lambertian assumption}, we decompose a face image into $5$ components: 3D geometry, common texture, identity texture, ambient light, and direct light, as shown in Figure~\ref{fig:lambertian}.
The 3D geometry is the underlying 3D face shape, the common texture is the albedo patterns shared by all the people, the identity texture is the albedo patterns peculiar to this face, the ambient light changes the face color globally, and the direct light generates shading.
We introduce how these components are obtained in Section~\ref{sec:motivation}.

\begin{figure}[htbp]
    \begin{center}
        \includegraphics[width=0.24\linewidth]{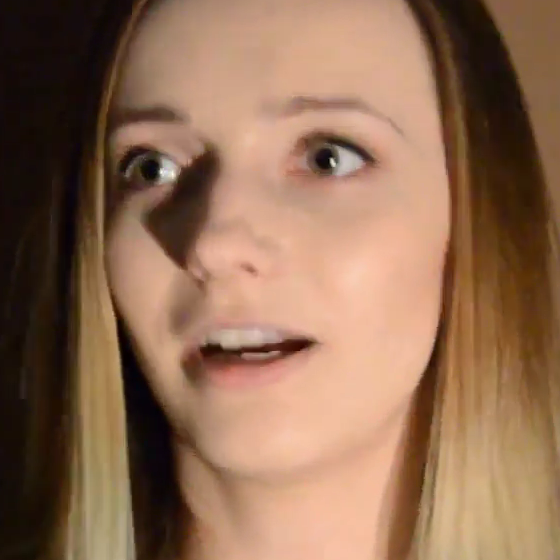}
        \includegraphics[width=0.24\linewidth]{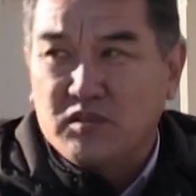}
        \includegraphics[width=0.24\linewidth]{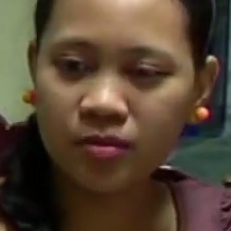}
        \includegraphics[width=0.24\linewidth]{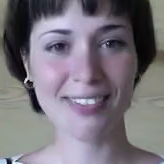}

        \includegraphics[width=0.24\linewidth]{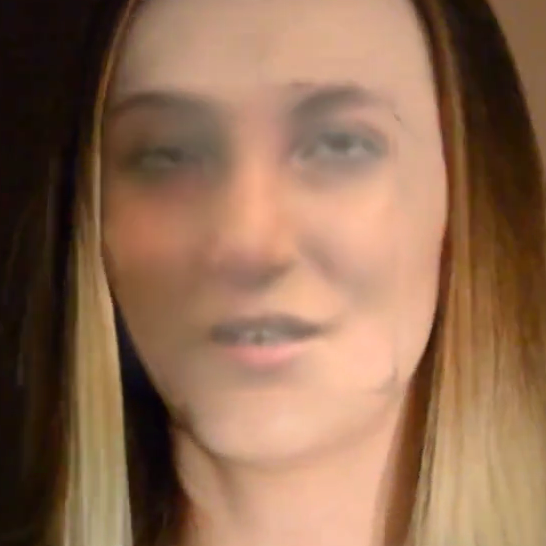}
        \includegraphics[width=0.24\linewidth]{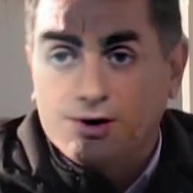}
        \includegraphics[width=0.24\linewidth]{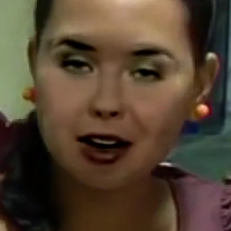}
        \includegraphics[width=0.24\linewidth]{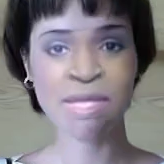}
    \end{center}
    \caption{
        Samples under strong direct light.
        The first row is the original faces, the second row is the corresponding fake samples, where evident inconsistency exists in the dim region.
    }
    \label{fig:fake-dirlight}
\end{figure}

Intuitively, the advanced manipulation methods can well reconstruct 3D geometry, common texture and ambient light since we merely see incompatible facial topology, non-face texture and weird skin color among the massive forged images.
Therefore, these three elements should be normalized.
On the contrary, we detect identity texture since it is hard to be simulated due to the rich variations across faces, leading to specific high-frequency artifacts.
Besides, we speculate that direct light is also a decisive forgery clue with the observation on large artifacts under intense direct light, shown in Figure~\ref{fig:fake-dirlight}.
By evaluating various compositions among different components, we find that the combination of \textbf{direct light} and \textbf{identity texture} is the best for forgery detection, which we call \textbf{facial detail}, as shown in Figure~\ref{fig:lambertian-f}.

When detecting forgery clues with neural networks, we consider the cooperation between face image and facial detail as a multi-modality task and propose a two-stream \textbf{F}orgery-\textbf{D}etection-with-\textbf{F}acial-\textbf{D}etail Net (FD$^{2}$Net).
To further highlight the discriminative region, we introduce a supervised Detail-guided Attention mechanism in the network, which employs the facial detail difference between real and fake faces as the objective.

In summary, our contributions are:
1) we introduce 3D decomposition into forgery detection and construct facial detail to amplify subtle artifacts.
2) A two-stream structure FD$^{2}$Net is proposed to fuse the clues from original images and facial details, where a supervised attention module is introduced to highlight the discriminative region.
3) Compared with the other state-of-the-art detection proposals, our method achieves remarkable elevation on both detection performance and generalization ability.

\section{Related Work}

\textbf{Digital face manipulation techniques}
There has been extensive research on face manipulation. Traditional methods require sophisticated editing tools, domain expertise, and time-consuming processes~\cite{tolosana2020deepfakes,verdoliva2020media,thies2015real,thies2016face2face,suwajanakorn2017synthesizing,kim2018deep}.
Recent deep learning (DL)-based methods, especially with GAN, have demonstrated their power on image synthesis, which promotes both face swapping and synthesis of entire fake images, making it more easy to be acquired by the public.
While the advanced manipulation techniques based on DL facilitate digital face manipulation remarkably, they exacerbate the difficulty for humans to distinguish manipulated faces from the genuine~\cite{rossler2019faceforensics++}.

\textbf{Manipulation detection method}
Facial forgery detection has attracted considerable attention recently, which stimulates massive study according to various forgery techniques~\cite{rahmouni2017distinguishing,carvalho2015illuminant,li2018ictu,guera2018deepfake}.
There is a large portion of methods discussing manipulation evidence among low-/high-level features.
Zhou \etal~\cite{zhou2017two} explore steganalysis features and propose to learn both tampering artifacts and local noise residual features.
Liu \etal~\cite{liu2020global} argue the effectiveness and robustness of global/large texture represented by the Gram matrix.
There are also methods of transferring images to the frequency domain to explore other forgery evidence~\cite{stuchi2017improving,qian2020thinking}.
However, previous texture-based methods extract facial features based on pixel-level images, \ie, merely concentrating on exploring manipulation trace among face appearance.

\textbf{Lighting-based detection}
There is also research focusing on detecting forgery evidence considering the lighting condition.
De Carvalho \etal~\cite{de2013exposing} propose to spot forgery evidence from the inconsistency among the 2D illuminant maps of various segments of the image.
Peng \etal~\cite{peng2016optimized} propose an optimized solution to estimate the 3D lighting environment.
However, these methods require comparison among at least two faces in one image, which is problematic in more common scenarios where only one face is in the image.

\section{Manipulation Detection with Facial Detail}

This paper regards face manipulation detection beyond a purely end-to-end binary classification problem.
We decompose a face image reversibly into several 3D descriptors, \ie, 3D shape, common texture, identity texture, ambient light and direct light, and explore how these descriptors contribute to the final label, investigating the best combination among these 3D descriptors for forgery detection.

\subsection{3D Decomposition}\label{sec:motivation}

In computer graphics, a face image is generated by:
\begin{equation}\label{eqn-img-render}
    \mathbf{I}_{syn} = \emph{Z-Buffer}(\mathbf{S}, \mathbf{C}),
\end{equation}
where $\mathbf{S}$ is the 3D face mesh, as shown in Figure~\ref{fig:lambertian-c}, and $\mathbf{C}$ is the RGB of each vertex in $\mathbf{S}$.
Under the \emph{Lambertian assumption}, the RGB of $i$th vertex is:
\begin{equation}\label{equ-lambertian}
    \begin{aligned}
        \mathbf{C}_{i} = \mathbf{Amb} * \mathbf{T}_{i} +  \langle \mathbf{n}_{i}, \mathbf{l} \rangle \cdot  \mathbf{Dir} * \mathbf{T}_{i},
    \end{aligned}
\end{equation}
where the facial texture $\mathbf{T}_{i}=[R_{i},G_{i},B_{i}]^{T}$ is the albedo of the $i$th vertex, $\mathbf{Amb}=\emph{diag}(R_{amb}, G_{amb}, B_{amb})$ is the color of the ambient light, as shown in Figure~\ref{fig:lambertian-b}, $\mathbf{n}_{i}$ is the vertex normal originating from the 3D mesh, $\mathbf{l}$ is the light direction, and $\mathbf{Dir}=\emph{diag}(R_{dir}, G_{dir}, B_{dir})$ is the color of the direct light, as shown in Figure~\ref{fig:lambertian-a}.

Then, we assume the facial texture $\mathbf{T}$ as the composition of common texture and identity texture, where the common texture $\mathbf{T}_{com}$ is the texture patterns shared by all the people, as shown in Figure~\ref{fig:lambertian-d}, and the identity texture $\mathbf{T}_{id}$ is the discriminative fine-grained texture containing one's identity information, as shown in Figure~\ref{fig:lambertian-e}.
In this paper, we model the common texture by the PCA texture model in Basel Face Model (BFM)~\cite{Paysan-AVSS-09}, and calculate the residual between $\mathbf{T}_{com}$ and $\mathbf{T}$ as the identity texture:
\begin{equation}\label{equ-texture}
    \begin{aligned}
        \mathbf{T} = \mathbf{\overline{T}} + \mathbf{B}\bm{\beta} + \mathbf{T}_{id},
    \end{aligned}
\end{equation}
where $\mathbf{\overline{T}}$ is the mean texture, $\mathbf{B}$ is the principle axes of the PCA texture model, and $\bm{\beta}$ is the common texture parameter.
Based on these models, any face images can be decomposed by a series of model parameters: $[\mathbf{S}, \mathbf{Amb}, \mathbf{Dir},\mathbf{\beta}, \mathbf{T}_{id}]$, which can be obtained by optimizing the following loss:
\begin{equation}\label{equ-description-get}
    \arg ~\min\limits_{\mathbf{S}, \mathbf{Amb}, \mathbf{Dir}, \bm{\beta}, \mathbf{T}_{id}}  ~ \| \mathbf{I} -  \mathbf{I}_{syn}(\mathbf{S}, \mathbf{Amb}, \mathbf{Dir}, \bm{\beta}, \mathbf{T}_{id}) \|,
\end{equation}
where $\mathbf{I}$ is the input face image.
After 3D decomposition, the following problems are whether each component contains forgery clues and how to combine them regarding the real/fake label.
Firstly, inspired by the previous discussions on high-frequency features beneath pixel-level texture~\cite{de2013exposing,li2020face}, we regard identity texture as a critical forgery clue and remove the topsoil facial texture, \ie, the ambient light and the common texture.
Secondly, by observing the fake samples under intensive direct light, as shown in Figure~\ref{fig:fake-dirlight}, we can consistently spot artifacts due to the large illumination difference between the source and target faces during manipulation.
Therefore, we suppose the existence of forgery clues in the direct light. Moreover, we emphasize the normalization of the 3D shape to make the detector concentrate on find-grained texture.

\begin{figure}[htbp]
    \centering
    \subfigure[]{
        \includegraphics[width=0.22\linewidth]{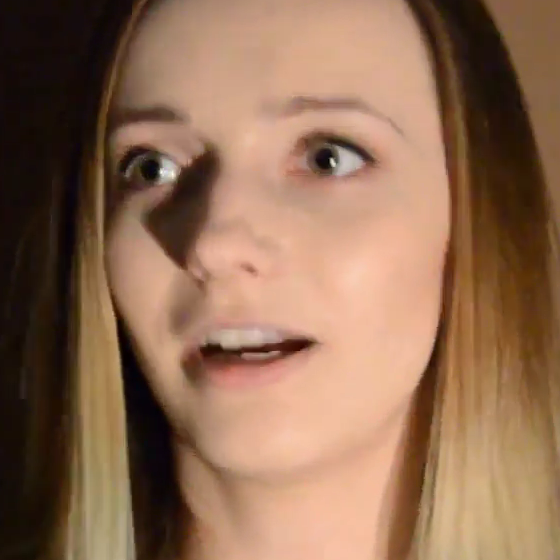}
        \label{fig:input-evaluate-a}
    }
    \subfigure[]{
        \includegraphics[width=0.22\linewidth]{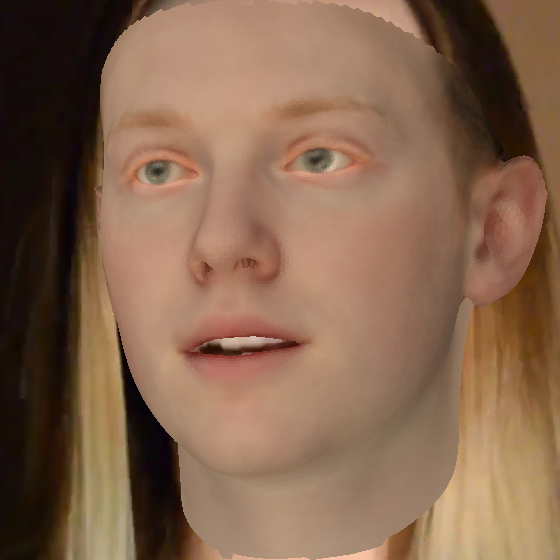}
        \label{fig:input-evaluate-b}
    }
    \subfigure[]{
        \includegraphics[width=0.22\linewidth]{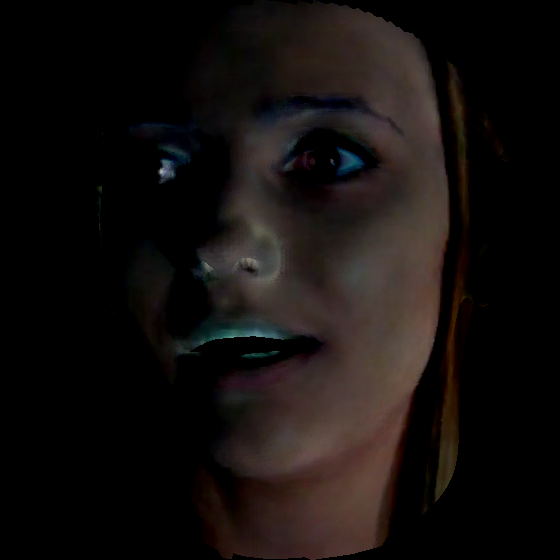}
        \label{fig:input-evaluate-c}
    }
    \subfigure[]{
        \includegraphics[width=0.22\linewidth]{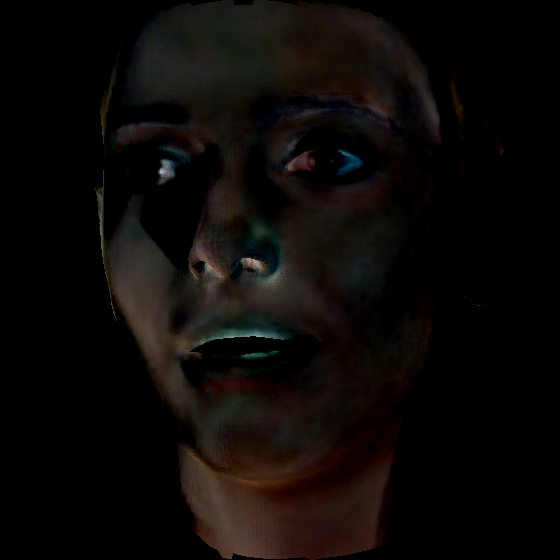}
        \label{fig:input-evaluate-d}
    }

    \subfigure[]{
        \includegraphics[width=0.22\linewidth]{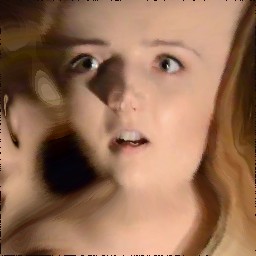}
        \label{fig:input-evaluate-e}
    }
    \subfigure[]{
        \includegraphics[width=0.22\linewidth]{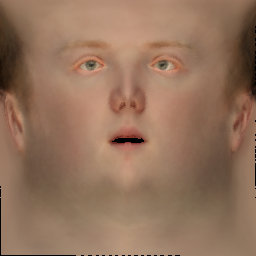}
        \label{fig:input-evaluate-f}
    }
    \subfigure[]{
        \includegraphics[width=0.22\linewidth]{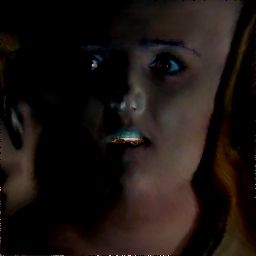}
        \label{fig:input-evaluate-g}
    }
    \subfigure[]{
        \includegraphics[width=0.22\linewidth]{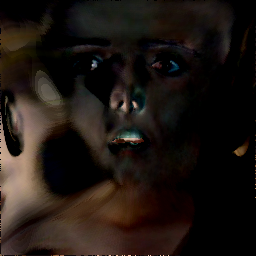}
        \label{fig:input-evaluate-h}
    }
    \caption{
        The $8$ inputs for forgery detection.
        (a) Face Image, (b) Ambient Light + Common Texture + 3D Shape, (c) Identity Texture + Direct Light + 3D Shape, (d) Identity Texture + 3D Shape, (e) Face Image w/o Shape, (f) Ambient Light + Common Texture, (g) Identity Texture + Direct Light, (h) Identity Texture.
    }
    \label{fig:input-evaluate}
\end{figure}

To verify the suppositions, we conduct a fast ablation study and propose $8$ inputs for forgery detection: \textbf{img}, \textbf{amb+ctex+shape}, \textbf{itex+dir+shape}, \textbf{itex+shape}, \textbf{img w/o shape}, \textbf{amb+ctex}, \textbf{itex+dir}, \textbf{itex}, where amb, dir, itex, ctex and shape are short for ambient light, direct light, identity texture, common texture and 3D shape, respectively, as shown in Figure~\ref{fig:input-evaluate}, where we warp the image to the UV space to discard the 3D shape.
We generate the $8$ inputs for all the samples in Faceforensics++~\cite{rossler2019faceforensics++} and train a VGG16 for evaluation.
The results are shown in Table~\ref{tab:input-evaluate}.

There are several noteworthy results in Table~\ref{tab:input-evaluate}.
Firstly, the poor performances of In-b and In-f indicate that the topsoil facial appearance, \ie, the ambient light and the common texture, are easy to be faked and have little forgery clues.
Secondly, by comparing In-c and In-g or comparing In-d and In-h, we find consistent improvements after warping the fine-grained appearance to the UV space. Therefore we suppose normalizing the 3D shape makes CNN concentrate on specific face regions and simplifies the detection task.
Thirdly, by comparing In-g and In-h, we find that direct light benefits the forgery detection remarkably.
Moreover, we find that many fake samples that In-c identifies but In-d does not are under intense light, verifying our assumption that current manipulation methods cannot simulate direct light properly.
In all, the 3D shape, ambient light, and common texture have few forgery patterns but contribute most of the pixel values, and they should be normalized, while the weak signals of direct light and identity texture should be highlighted due to the embedded critical clues.
In the following implementation, we use \textbf{facial trend} to represent the group of 3D shape, ambient light and common texture, and name the combination of direct light and identity texture as the \textbf{facial detail}.

\begin{table}[htbp]
    \begin{center}
        \begin{tabular}{c|ccccc|ccc}
            \hline\hline
            input & shape      & amb        & dir        & ctex       & itex       & AUC            \\
            \hline\hline
            In-a  & \checkmark & \checkmark & \checkmark & \checkmark & \checkmark & 99.13          \\
            \hline
            In-b  & \checkmark & \checkmark &            & \checkmark &            & 50.00          \\
            \hline
            In-c  & \checkmark &            & \checkmark &            & \checkmark & 99.29          \\
            \hline
            In-d  & \checkmark &            &            &            & \checkmark & 99.14          \\
            \hline\hline
            In-e  &            & \checkmark & \checkmark & \checkmark & \checkmark & 98.93          \\
            \hline
            In-f  &            & \checkmark &            & \checkmark &            & 50.00          \\
            \hline
            In-g  &            &            & \checkmark &            & \checkmark & \textbf{99.56} \\
            \hline
            In-h  &            &            &            &            & \checkmark & 99.27          \\
            \hline\hline
        \end{tabular}
    \end{center}
    \caption{
        The AUC performance (\%) on \emph{Faceforensics++} (FFpp)~\cite{rossler2019faceforensics++}. The inputs are the compositions of $5$ components, including: 3D shape (shape), ambient light (amb), direct light (dir), common texture (ctex) and identity texture (itex).
        The examples of In-a to In-h are shown in Figure~\ref{fig:input-evaluate}.
        The best results are highlighted.
    }
    \label{tab:input-evaluate}
\end{table}

\subsection{Facial Detail Generation}
Based on the analysis in 3D decomposition, we aim to normalize facial trend (the combination of 3D shape, ambient light and common texture) and highlight facial detail (the combination of direct light and identity texture).
A trivial method is optimizing all the parameters together as Eqn.~\ref{equ-description-get} in an analysis-by-synthesis~\cite{1_blanz2003face} manner, but it costs too much computation.
Thus, we propose an approximation to expedite the generation of facial detail for fast inference.
We begin with the real-time generation of the 3D shape $\mathbf{S}$ by the state-of-the-art 3DDFA~\cite{zhu2017face,3ddfa_cleardusk}.
Then, we keep the 3D shape $\mathbf{S}$ and get the ambient and direct light by the spherical harmonic reflectance on the mean texture.

The spherical harmonics~\cite{zhang2006face}:
\begin{equation}
    \mathbf{H} = [\mathbf{h}_{1}, \mathbf{h}_{2}, \ldots, \mathbf{h}_{9}]
\end{equation}
are a set of functions that form an orthonormal basis to represent the brightness changes due to illuminations:
\begin{equation}\label{equ-lambertian}
    \begin{aligned}
        \mathbf{h}_{1} & = \frac{1}{\sqrt{4\pi}},                                                                                                                           \\
        \mathbf{h}_{2} & = \sqrt{\frac{3}{4\pi}}\mathbf{n}_x, ~~~\mathbf{h}_{3} = \sqrt{\frac{3}{4\pi}}\mathbf{n}_y, ~~~\mathbf{h}_{4} = \sqrt{\frac{3}{4\pi}}\mathbf{n}_z, \\
        \mathbf{h}_{5} & = \frac{1}{2}\sqrt{\frac{5}{4\pi}}(2\mathbf{n}_{z^2} - \mathbf{n}_{x^2} - \mathbf{n}_{y^2}),                                                       \\
        \mathbf{h}_{6} & = 3\sqrt{\frac{5}{12\pi}}\mathbf{n}_{yz}, ~~~\mathbf{h}_{7} = 3\sqrt{\frac{5}{12\pi}}\mathbf{n}_{xz},                                              \\
        \mathbf{h}_{8} & = 3\sqrt{\frac{5}{12\pi}}\mathbf{n}_{xy}, ~~~\mathbf{h}_{9} = \frac{3}{2}\sqrt{\frac{5}{12\pi}}(\mathbf{n}_{x^2}-\mathbf{n}_{y^2})
    \end{aligned}
\end{equation}
where $\mathbf{n}_{x}, \mathbf{n}_{y}, \mathbf{n}_{z}$ are the $x,y,z$ of the vertex normals computed by the 3D mesh $\mathbf{S}$, and we use $\mathbf{n}_{x^2}$ to denote a vector such that $\mathbf{n}_{x^2,i} = \mathbf{n}_{x,i}\mathbf{n}_{x,i}$ for the $i$th vertex and define $\mathbf{n}_{y^2}$, $\mathbf{n}_{z^2}$, $\mathbf{n}_{xz}$, $\mathbf{n}_{yz}$, and $\mathbf{n}_{xy}$ similarly.
With this set of basis, the face appearance under arbitrary illumination can be represented by the linear combination $(\mathbf{H} \bm{\gamma}) \cdot \mathbf{T}$, where $\mathbf{T}$ is the facial texture (vertex albedo), $\bm{\gamma}=[\gamma_{1}, \gamma_{2}, \ldots, \gamma_{9}]$ is the $9$-dimensional reflectance parameters and $\cdot$ is the dot product. We consider $\gamma_{1}\cdot\mathbf{h}_{1}$ as the ambient light and $[\gamma_{2} \cdot \mathbf{h}_{2}, \ldots, \gamma_{9} \cdot \mathbf{h}_{9}]$ as the direct light.

In our implementation, we degrade the $\mathbf{T}$ to the mean texture $\mathbf{\overline{T}}$ for fast inference and get $\bm{\gamma}$ from the least squares solution of the following equation:
\begin{equation}\label{equ-illum-get}
    \mathbf{I}(\mathbf{S})=(\mathbf{H} \bm{\gamma}) \cdot \mathbf{\overline{T}},
\end{equation}
where $\mathbf{I}(\mathbf{S})$ are the pixels at vertex positions. Based on the harmonic reflectance parameters, we further get the common texture by the following linear equation:
\begin{equation}\label{equ-ctex-get}
    \mathbf{I}(\mathbf{S})=(\mathbf{H} \bm{\gamma}) .* (\mathbf{\overline{T}} + \mathbf{B}\bm{\beta}),
\end{equation}
where $\mathbf{\overline{T}}$ and $\mathbf{B}$ are from the PCA texture model, $\bm{\gamma}$ is the reflectance parameters estimated in Eqn.~\ref{equ-illum-get} and $\bm{\beta}$ is the common texture parameters.
Finally, we obtain the facial detail by:
\begin{equation}\label{equ-facial-detail-get}
    \mathbf{FD} = UV(\mathbf{I} - (\mathbf{h}_{1} \gamma_{1}) .* (\mathbf{\overline{T}} + \mathbf{B}\bm{\beta}), \mathbf{S}),
\end{equation}
where $\mathbf{FD}$ is the facial detail and $UV(\mathbf{I},\mathbf{S})$ is the UV warping that transfers image pixels in $\mathbf{I}$ to the UV space by the constraints of 3D mesh $\mathbf{S}$.
We suppose that the facial detail highlights the forgery patterns for the forged image, making it more suitable for the input of manipulation detection neural networks.

\section{FD$^{2}$Net}

\begin{figure*}[htbp]
    \subfigure{
        \label{fig:full_net-a}}
    \subfigure{
        \label{fig:full_net-b}}
    \subfigure{
        \label{fig:full_net-c}}
    \begin{center}
        \includegraphics[width=0.8\linewidth]{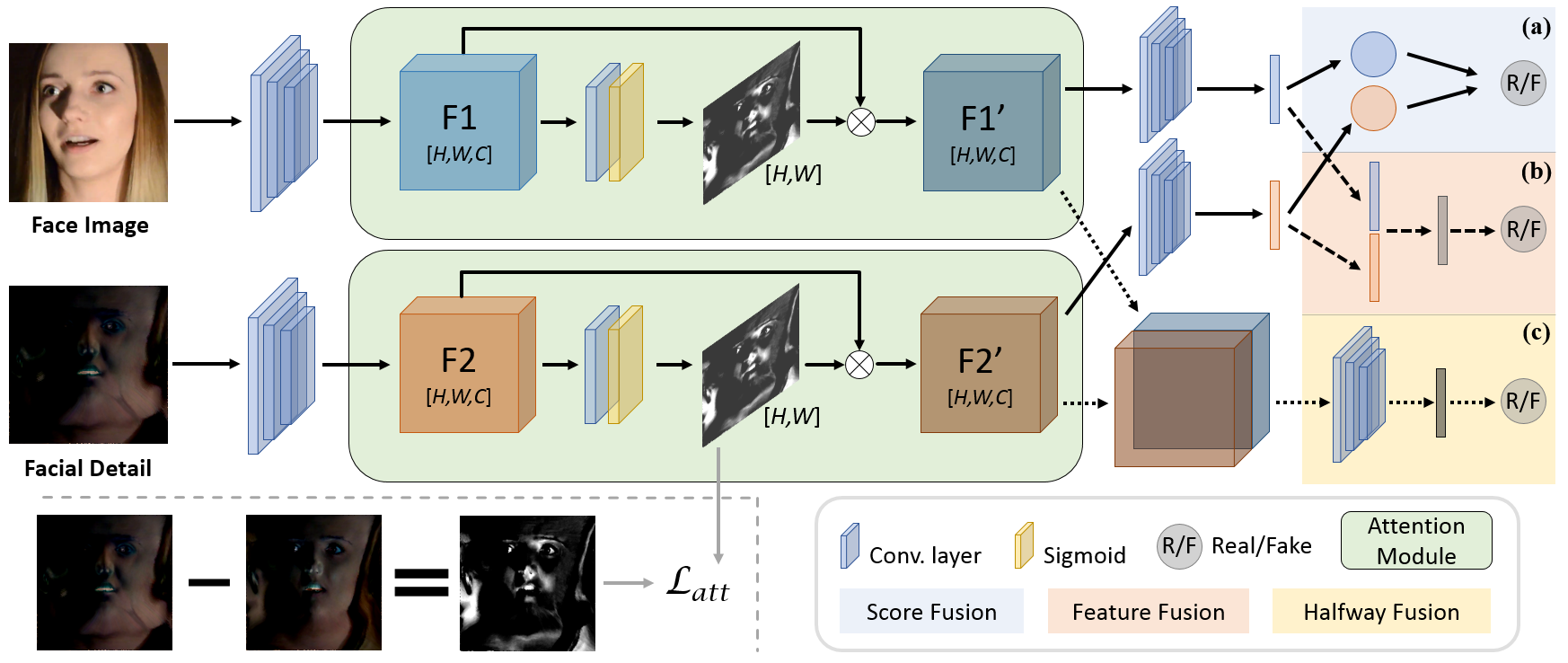}
    \end{center}
    \caption{
        The overview of the \textbf{F}orgery-\textbf{D}etection-with-\textbf{F}acial-\textbf{D}etail Net (FD$^{2}$Net).
        We introduce a two-stream structure to combine the clues from original images and facial details. The results of the two streams are fused by three methods: (a) score fusion (SF), (b) feature fusion (FF) and (c) halfway fusion (HF). Besides, we insert a detail-guided attention module, which is supervised by the facial detail difference, in the middle of the backbone network.
    }
    \label{fig:full_net}
\end{figure*}

Our \textbf{F}orgery-\textbf{D}etection-with-\textbf{F}acial-\textbf{D}etail Net (FD$^{2}$Net) is briefly presented in Figure~\ref{fig:full_net}.
We adopt the state-of-the-art XceptionNet~\cite{chollet2017xception} as the backbone and then merge the two-stream structure and the attention mechanism into it to explore the potential enhancement assisted by the correlation among the multi-modality data and the location of forgery clues.

\subsection{Multi-modality Fusion with Two-stream} \label{sec:2str}
Although the facial detail highlights the critical clues in fine-grained texture and shading, it may warp specific forgery patterns and miss external face regions.
Therefore, we consider facial detail and original face image as complementary clues and characterize forgery detection as a multi-modality task, regarding facial detail and pixel-level face image as two different modalities.
To be specific, we adopt a two-stream architecture to study the combination of these two modalities, where each stream is equipped with the XceptionNet~\cite{chollet2017xception} to detect face image and facial detail separately.
The classifier performs cooperated decisions at the end regarding the joint representations of the two streams.

We evaluate three ways of fusing the representations of the two modalities, as illustrated in Figure~\ref{fig:full_net}.
Firstly, we implement the score fusion (SF), which performs real/fake classification in each stream and adding their confidences as the final score.
The result is real if the score is larger than $1$ and fake otherwise, shown in Figure~\ref{fig:full_net-a}.
Secondly, we implement the feature fusion (FF), where each stream ends with a fully-connected (FC) layer, and their features are concatenated to a one-dimension vector and transferred by another FC layer to make the final decision, shown in Figure~\ref{fig:full_net-b}.
Finally, we implement the halfway fusion (HF) to concatenate the intermediate 2D feature maps for further single-stream processing, shown in Figure~\ref{fig:full_net-c}.
To be specific, we convolve the two inputs by the first half backbone, \ie, before the 7th block of XceptionNet~\cite{chollet2017xception}, and stack their 2D outputs as the feature map afterward.
Then we adopt the last half XceptionNet to process the stacked feature in a single stream to make the final decision. In our experiments, we find halfway fusion performs the best with smaller parameter size, which may benefit from preserving spatial information when fusing the local features.

\begin{table*}[htbp]
    \begin{center}
        \begin{tabular}{c|ccc|ccc|ccc}
            \hline\hline
            \multirow{2}{*}{Structure} & \multicolumn{3}{c|}{FFpp} & \multicolumn{3}{c|}{DFD} & \multicolumn{3}{c}{DFDC}                                                                                                       \\
            \cline{2-10}
                                       & AP                        & AUC                      & EER                      & AP             & AUC            & EER            & AP             & AUC            & EER            \\
            \hline\hline
            Img                        & 99.44                     & 99.31                    & 5.39                     & 88.07          & 65.57          & 38.38          & 85.60          & 62.17          & 39.99          \\
            \hline
            Detail                     & 99.40                     & 99.12                    & 5.51                     & 87.24          & 64.29          & 40.87          & 85.02          & 61.80          & 40.37          \\
            \hline
            Img ($\times$ 2)           & 99.67                     & 99.38                    & 5.37                     & 89.45          & 74.14          & 34.07          & 86.70          & 63.22          & 38.77          \\
            \hline\hline
            Img+Detail (SF)            & 97.84                     & 92.91                    & 11.07                    & 83.71          & 72.82          & 36.44          & 81.91          & 62.10          & 47.32          \\
            \hline
            Img+Detail (FF)            & \textbf{99.72}            & \textbf{99.45}           & \textbf{5.31}            & 89.56          & 78.55          & 26.80          & 87.16          & 65.36          & 36.17          \\
            \hline
            Img+Detail (HF)            & 99.42                     & 98.73                    & 5.63                     & \textbf{89.61} & \textbf{78.65} & \textbf{26.03} & \textbf{87.31} & \textbf{66.09} & \textbf{35.46} \\
            \hline\hline
        \end{tabular}
    \end{center}
    \caption{
        Test results (\%) of the two-stream FD$^{2}$Net and its variants on FFpp, DFD and DFDC.
        The ``Img'' is the stream detecting original images only.
        The ``Detail'' is the stream detecting facial details only.
        The ``Img ($\times$ 2)'' is the one-stream network on original images but having the same parameter size as the two-stream structure.
        The SF, FF and HF refer to score fusion, feature fusion and halfway fusion, respectively.
        The best results are highlighted.
    }
    \label{table:result_ffpp_2str}
\end{table*}

\subsection{Detail-guided Attention}
Extensive tasks have adopted the attention mechanism to enhance forgery detection performance~\cite{dang2020detection}.
The embedded attention module exploits more distinguishing characteristics by positioning the plausible manipulated region, and also strengthens the explainability of the classifiers~\cite{dang2020detection,bonettini2020video}.
Unlike the previous methods, which either need the ground truth manipulated regions or adaptively learn the attention map by the real/fake labels~\cite{dang2020detection}, we supervise our attention map by the facial detail difference between fake and real faces.

Generally, an attention map $\mathbf{M}_{att}$ is constructed from an intermediate feature map $\mathbf{F}$ by a small regression network $\mathbf{M}_{att}=\mathcal{N}(\mathbf{F}, \theta_{att})$ with $\theta_{att}$ as its parameters.
Then the intermediate feature is refined by the attention map $\mathbf{F}' = \mathbf{F} \bigotimes Sigmoid(\mathbf{M}_{att})$, where $\bigotimes$ denotes element-wise multiplication.
In this work, we propose a novel approach to train the attention network $\theta_{att}$.
When constructing the batches during network training, two images are selected for each sample, one real $\mathbf{I}_{real}$ and one fake $\mathbf{I}_{fake}$.
The absolute of the grayscale facial detail difference is taken as a weak supervision of the attention module:
\begin{equation}\label{equ-ctex-get}
    \mathcal{L}_{att} =  \|  ~\mathcal{N}(\mathbf{F}, \theta_{att}) - | FD(\mathbf{I}_{real}) - FD({\mathbf{I}_{fake}})|~ \|,
\end{equation}
where $FD(\cdot)$ is facial detail extraction.
Then the total loss is:
\begin{equation}\label{equ-ctex-get}
    \mathcal{L} =  \mathcal{L}_{cls} + \lambda_{att}\mathcal{L}_{att},
\end{equation}
where $\lambda_{att}$ is the weight of attention loss and $\mathcal{L}_{cls}$ is the cross entropy loss performing real/fake classification.

\section{Experiments}
\label{experiments}
In this section, we introduce the datasets, experiment setups, extensive experiment results of the ablation studies, and comparison with previous works in sequence.

\noindent \textbf{Training Dataset}.
\emph{Faceforensics++} (FFpp)~\cite{rossler2019faceforensics++} is a benchmark dataset released recently for facilitating evaluation among facial manipulation detection methods.
There are $1k$ original video sequences, in which $720$, $140$, $140$ videos are used for training, validation and testing, respectively.
These original videos are manipulated by four state-of-the-art face manipulation methods, \ie, \emph{DeepFakes} (DF)~\cite{deepfake2020github}, \emph{Face2Face} (F2F)~\cite{thies2016face2face}, \emph{FaceSwap} (FS)~\cite{faceswap2020github}, and \emph{NeuralTextures} (NT)~\cite{thies2019deferred}.
Besides, the raw video sequences are degraded with different compression rate ($0$, $23$, $40$) to simulate the real situation~\cite{rossler2019faceforensics++}.
We select the HQ version (c$23$) of FFpp, considering the extensive post-processing imposed on the original data before they go public, and sample $100$ frames for each video in the experiments.

\noindent \textbf{Test Datasets}.
We adopt the following datasets for performance and generalization evaluation.
$1)$ the testing set of FFpp as described above.
$2)$ \emph{The DeepFake Detection dataset} (DFD)~\cite{dfd2019blog} containing hundreds of original data and thousands of manipulated data, released by Google for promoting research on synthetic video detection.
$3)$ \emph{Deepfake detection challenge dataset} (DFDC)~\cite{DFDC2020} containing over $100$k video sequences captured with over $3$k paid actors and manipulated videos covering Deepfake, GAN-based, and non-learned methods, released recently for the corresponding Kaggle competition\footnote{https://www.kaggle.com/c/deepfake-detection-challenge} by Facebook AI.

\noindent \textbf{Implementation Details}.
For the facial detail generation, we construct the 3D face shape by 3DDFA~\cite{zhu2017face,3ddfa_cleardusk}, perform UV warping by the UV map in~\cite{16_feng2018joint}, and acquire the common texture by fitting the PCA texture model in Basel Face Model (BFM)~\cite{Paysan-AVSS-09}.
For the neural network, we introduce XceptionNet~\cite{chollet2017xception} as the backbone and fuse the feature map after the 4th block of the middle row of XceptionNet when implementing the halfway fusion structure.
The Adam optimizer is utilized for training with weight decay equals to $5 \times 10^{-4}$, $\beta_1 = 0.9$, $\beta_2 = 0.999$ and batch size set to $32$.
The initial learning rate is $10^{-4}$, then changed to $5\times10^{-5}$ at epoch $15$, to $5\times10^{-6}$ at epoch $23$, to $10^{-6}$ at epoch $28$, and to $5\times10^{-7}$ for the rest from epoch $32$.
An early-stop module controls the training process's end, terminating the training if the loss on the validation set does not fall for $7$ consecutive epochs.
In our implementation, the total epoch is about $25$.
Besides, the $\lambda_{att}$ in the loss function is set to $1$.

\subsection{Ablation Studies}

\begin{table*}[htbp]
    \begin{center}
        \begin{tabular}{cc|ccc|ccc|ccc}
            \hline\hline
            \multicolumn{2}{c|}{Attention on Stream} & \multicolumn{3}{c|}{FFpp} & \multicolumn{3}{c|}{DFD} & \multicolumn{3}{c}{DFDC}                                                                                                                       \\
            \hline
            Img Attention                            & Detail Attention          & AP                       & AUC                      & EER           & AP             & AUC            & EER            & AP             & AUC            & EER            \\
            \hline
                                                     &                           & 99.42                    & 98.73                    & 5.63          & 89.61          & 78.65          & 26.03          & 87.31          & 66.09          & 35.46          \\
            \checkmark                               &                           & 99.45                    & 98.73                    & 5.62          & 89.65          & 78.71          & 25.94          & 87.58          & 66.51          & 35.33          \\
                                                     & \checkmark                & 99.47                    & 98.74                    & 5.51          & 89.37          & 78.69          & 26.01          & 87.56          & 66.48          & 35.33          \\
            \checkmark                               & \checkmark                & \textbf{99.48}           & \textbf{98.76}           & \textbf{5.59} & \textbf{89.84} & \textbf{79.08} & \textbf{25.18} & \textbf{87.93} & \textbf{67.70} & \textbf{34.91} \\
            \checkmark(unsupervised)                 & \checkmark (unsupervised) & 99.44                    & 98.68                    & 5.88          & 88.55          & 78.37          & 27.46          & 87.02          & 65.46          & 37.23          \\
            \hline
            \hline
        \end{tabular}
    \end{center}
    \caption{
        The ablation study results (\%) on the Detail-guided Attention module in FD$^{2}$Net.
        The ``Img Attention'' and ``Detail Attention'' refer to the attention module on the image stream and detail stream, respectively.
        We also explore the performance without the supervised signal by the facial detail difference and present the last row results with ``unsupervised'' in the bracket.
        The best results are highlighted.
    }
    \label{table:result_attention}
\end{table*}

\begin{table*}[htbp]
    \begin{center}
        \begin{tabular}{c|cc|ccc|ccc|ccc}
            \hline\hline
            S$1$       & \multicolumn{2}{c|}{S$2$} & \multicolumn{3}{c|}{FFpp} & \multicolumn{3}{c|}{DFD} & \multicolumn{3}{c}{DFDC}                                                                                                                       \\
            \hline
            Image      & Tex Norm                  & Shape Norm                & AP                       & AUC                      & EER           & AP             & AUC            & EER            & AP             & AUC            & EER            \\
            \hline
            \checkmark &                           &                           & 99.46                    & 99.47                    & 4.48          & 88.14          & 72.51          & 36.77          & 85.64          & 62.28          & 39.56          \\
            \hline
            \checkmark &                           & \checkmark                & 99.57                    & 99.59                    & 4.30          & 84.06          & 76.09          & 29.11          & 86.13          & 64.43          & 38.40          \\
            \checkmark & \checkmark                &                           & \textbf{99.61}           & \textbf{99.68}           & \textbf{4.28} & 85.10          & 76.84          & 27.08          & 87.73          & 66.01          & 38.32          \\
            \checkmark & \checkmark                & \checkmark                & 99.48                    & 98.76                    & 5.59          & \textbf{89.84} & \textbf{79.08} & \textbf{25.18} & \textbf{87.93} & \textbf{67.70} & \textbf{34.91} \\
            \hline
            \hline
        \end{tabular}
    \end{center}
    \caption{
        The ablation study result (\%) of facial detail in FD$^{2}$Net.
        The S$1$ and S$2$ refer to the first and the second stream in the network.
        The ``Tex Norm'' refers to texture normalization and the ``Shape Norm'' refers to shape normalization.
        The best results are highlighted.
    }
    \label{table:result_ffpp}
\end{table*}

\subsubsection{Analysis of the Two-stream Network}

We regard face image and facial detail as two complementary modalities and implement a two-stream network to fuse their clues.
To evaluate each modality's performance and the best fusion manner, we quantitatively evaluate FD$^{2}$Net in different variants: one-stream with original images only, one-stream with facial details only, and two-stream fused by score-fusion, feature-fusion, and halfway-fusion, respectively.
We do not adopt the attention module here.
The results are listed in Table~\ref{table:result_ffpp_2str}.

Firstly, the one-stream structure considering face images or facial details achieves similar results, but worse than that of the two-stream structure, especially in cross-data evaluation.
Secondly, the deteriorated performance of the score fusion indicates the potential sophistication of multi-modality clues fusion and the inflexibility of the hand-crafted decision function.
Nevertheless, the performance on FFpp, DFD, and DFDC become jointly better with the feature fusion, validating the complementarity of the two modalities.
Finally, the halfway fusion further promotes the results on DFD and DFDC by the local fusion manner, which makes the fused features correspond to similar receptive fields.
We also find that the two-stream structure outperforms the one-stream with double parameters, ruling out the benefit from a larger parameter size on the performance improvement.

\subsubsection{Analysis of the Detail-guided Attention}

To highlight the plausible manipulated region, we introduce the detail-guided attention module between the fourth and fifth block of the middle flow of the XceptionNet.
Based on the two-stream network with halfway fusion, we evaluate some of the network variants by separately deploying the attention module on each stream and exploring whether the supervised signal improves its effectiveness in further discussion.
Results in Table~\ref{table:result_attention} demonstrate the improvements of XceptionNet on all datasets with the additional attention module, either implemented on the image stream or the detail stream. The network with attention modules on both streams achieves the best performance.
Besides, we train the attention module indirectly from the real/fake labels, ignoring the supervised signal by the facial detail difference, and observe a performance drop, indicating the effectiveness of the supervised signals on the forgery detection.

\subsubsection{Ablation Study of Facial Detail in FD$^{2}$Net}

Although Table~\ref{tab:input-evaluate} has performed an extensive ablation study on each 3D component, we further evaluate facial detail when acting as a complementary clue in the two-stream network.
In this section, we decompose the effectiveness of facial detail into shape normalization and texture normalization, where shape normalization refers to warping the facial pixels to the UV space, \eg, Figure~\ref{fig:input-evaluate-e}, and texture normalization refers to decomposing and removing ambient light and common texture in the pixel values, \eg, Figure~\ref{fig:input-evaluate-c}.
We adopt the two-stream network with both halfway fusion and supervised attention module and present the results in Table~\ref{table:result_ffpp}.

The first row is a one-stream network which directly detects forgery from original face images without using any facial detail information.
Adding the second stream can promote the AUC scores compared with the primary one-stream structure, either implementing shape or texture normalization.
Furthermore, the introduction of the facial detail, \ie, adopting the combination of shape and texture normalization, helps the detector achieve the best performance.
These progressive improvements validate that the proposed facial detail contributes to the forgery detection and complements the original image.

\subsection{Comparison with other methods}

\begin{table*}[htbp]
    \begin{center}
        \begin{tabular}{c|c|ccc|ccc}
            \hline\hline
            \multirow{2}{*}{Model}                     & \multirow{2}{*}{Training dataset} & \multicolumn{3}{c|}{DFD (HQ)} & \multicolumn{3}{c}{DFDC}                                                                     \\
            \cline{3-8}
                                                       &                                   & AP                            & AUC                      & EER            & AP             & AUC            & EER            \\
            \hline
            Xception~\cite{rossler2019faceforensics++} & FFpp                        & 88.07                         & 65.57                    & 38.38          & 85.60          & 62.17          & 39.99          \\
            \hline
            \tabincell{c}{EfficientNetB4                                                                                                                                                                                  \\ Ensemble~\cite{bonettini2020video}} &   FFpp                       & 89.35                         & 72.82                    & 34.86       & 85.71       & 63.03 & 38.86       \\
            \hline\hline
            FD$^{2}$Net                                & FFpp              & \textbf{89.84}                & \textbf{79.08}           & \textbf{25.18} & \textbf{87.93} & \textbf{67.70} & \textbf{34.91} \\
            \hline
            \hline
        \end{tabular}
    \end{center}
    \caption{
        Performance (\%) comparison among previous state-of-the-art methods on the unseen dataset, DFD (HQ) and DFDC.
        The best results are highlighted.
    }
    \label{table:result_other_dfd_dfdc}
\end{table*}

\begin{table}[htbp]
    \begin{center}
        \begin{tabular}{c|c|cc}
            \hline\hline
            \multirow{2}{*}{Model}                  & \multirow{2}{*}{Training data} & \multicolumn{2}{c}{Acc}                  \\
            \cline{3-4}
                                                    &                                & F2F                     & FS             \\
            \hline
            MesoInception4~\cite{afchar2018mesonet} & \multirow{7}{*}{F2F}           & 84.56                   & 56.71          \\
            VA-LogReg~\cite{matern2019exploiting}   &                                & 83.62                   & 59.45          \\
            LAE~\cite{du2019towards}                &                                & 90.34                   & 62.51          \\
            Multi-task~\cite{nguyen2019multi}       &                                & 91.27                   & 55.04          \\
            Face X-ray~\cite{li2020face}            &                                & 97.73                   & 85.69          \\
            Xception + HP Filter                    &                                & 97.98                   & 57.46          \\
            FD$^{2}$Net                             &                                & \textbf{98.22}          & \textbf{86.54} \\
            \hline\hline
        \end{tabular}
    \end{center}
    \caption{
        Detection accuracy comparison (\%) with previous methods on F2F and FS in FFpp.
        We adopt the HQ (c$23$) version data from FFpp to discuss the robustness on the unseen manipulation technique.
        The best results are highlighted.
    }
    \label{table:result_other_f2f_fs}
\end{table}

Some previous works~\cite{khodabakhsh2018fake,du2019towards,nguyen2019multi,li2020face} indicate the potential generalization failure when detecting unseen manipulation methods or datasets.
In this section, we compare our method with previous state-of-the-art methods to explore our  performance in both scenarios.

\textbf{Cross-data Evaluation}.
Following Khodabakhsh \etal~\cite{khodabakhsh2018fake}, we quantitatively analyze the generalization ability on unseen data and compare it with other methods, including the primary XceptionNet detection method~\cite{rossler2019faceforensics++} and the ensemble of EfficientNet's variants~\cite{bonettini2020video}.
We train the model on FFpp, test it on DFD (HQ) and DFDC following the ablation study's configuration, and list the results in Table~\ref{table:result_other_dfd_dfdc}.
The improvement in the generalization on unseen data demonstrates that the additional facial detail enables the detection model to effectively extract more discriminative and general features from fake images, even from a different distribution of the training dataset.
It is worth noting that the extraction of facial detail is independent of any forgery data, making it the probable reason for better generalization.

\textbf{Evaluations on Different Manipulation Methods}.
Following Li \etal~\cite{li2020face}, we evaluate the robustness of our method on the unseen manipulation methods and compare the performance with previous methods.
We introduce the data manipulated by different methods, \ie, Face2Face (F2F) and FaceSwap (FS) under the low compression (c23) from FFpp, and train our approach on F2F and test it on both F2F and FS, taking the correct prediction accuracy as the evaluation metric.
The results are listed in Table~\ref{table:result_other_f2f_fs}.
The proposed method achieves $98.22\%$ on F2F and $86.54\%$ on FS, with a significant improvement compared to the current state-of-the-art.
The improvements mainly benefit from the highlighted clues extracted from the facial detail and the plausible forged regions indicated by the attention map.
In particular, some compared methods also consider complementary information from various modalities.
Nguyen \etal~\cite{nguyen2019multi} propose sharing knowledge learned simultaneously from images and videos to enhance the performance of the detection on both data.
Li \etal~\cite{li2020face} explore the estimation of the blending boundary directly from face image to discover the possibility of decomposing an image into the mixture of two images from different sources.
Besides, we also include signal decomposition methods in frequency domains, introducing the three base high-pass filters in the \emph{FAD} stream in~\cite{qian2020thinking} to primary XceptionNet (Xception + HP Filter).
Unlike these methods, the proposed FD$^{2}$Net strips the ambient lighting and the common appearance with 3D decomposition, exploiting the personalized ambient-free facial detail to extract more robust discriminative features.

\section{Conclusion}

This paper proposes a novel face forgery detection method by the 3D decomposition of the face image.
By disentangling the face image into 3D shape, common texture, identity texture, ambient light, and direct light, we find critical forgery clues in the direct light and the identity texture.
To utilize this observation, we propose the facial detail, which is constructed by warping image pixels to the UV space and removing the topsoil facial texture, to highlight the subtle forgery patterns.
The clues in the facial detail and the original image are fused by a two-stream network FD$^{2}$Net for the final real or fake classification.
Meanwhile, an attention module supervised by the facial detail is proposed to highlight the plausible manipulated region.
Extensive experiments demonstrate the effectiveness and generalization of the proposed FD$^{2}$Net on the FaceForencis++ dataset.
In general, our work provides a novel direction to find the forgery clues by analyzing how an image is generated in physics, following the analysis-by-synthesis idea.

    {\small
        \bibliographystyle{ieee_fullname}
        \bibliography{mybib}
    }
\end{document}